%% file: main.tex
\documentclass[10pt,twocolumn,letterpaper]{article}

\usepackage{cvpr}              
\input{preamble}
\definecolor{cvprblue}{rgb}{0.21,0.49,0.74}
\usepackage[pagebackref,breaklinks,colorlinks,allcolors=cvprblue]{hyperref}

\usepackage{multirow}
\usepackage{makecell}
\definecolor{firstplace}{RGB}{255,179,179}   
\definecolor{secondplace}{RGB}{255,217,179} 
\definecolor{thirdplace}{RGB}{255,255,180}   

\def\confName{CVPR}
\def\confYear{2026}

\title{

Intrinsic Geometry-Appearance Consistency Optimization for \\
Sparse-View Gaussian Splatting

}

\author{Kaiqiang Xiong \textsuperscript{1,2} \quad Rui Peng \textsuperscript{4} \quad Jiahao Wu \textsuperscript{1,2} \quad Zhanke Wang \textsuperscript{1} \\
Jie Liang \textsuperscript{1,2} \quad Xiaoyun Zheng \textsuperscript{2} \quad Feng Gao \textsuperscript{5} \quad Ronggang Wang \textsuperscript{1,2,3}\\
\textsuperscript{1}Guangdong Provincial Key Laboratory of Ultra High Definition Immersive Media Technology,\\ Shenzhen Graduate School, Peking University \\
\textsuperscript{2}Peng Cheng Laboratory \quad \textsuperscript{3}Migu Culture Technology Co., Ltd \\ \textsuperscript{4} Alibaba Group \quad \textsuperscript{5}School of Arts, Peking University \\
{\tt\small xiongkaiqiang@stu.pku.edu.cn \quad rgwang@pkusz.edu.cn} \\
{ \url{https://KaiqiangXiong.github.io/ICO-GS/}}}

\begin{document}

\makeatletter
\let\@oldmaketitle\@maketitle
\renewcommand{\@maketitle}{\@oldmaketitle

\resizebox{1.0\linewidth}{!}{
\includegraphics[trim={0cm 0cm 0cm 0cm},clip, width=1\linewidth]{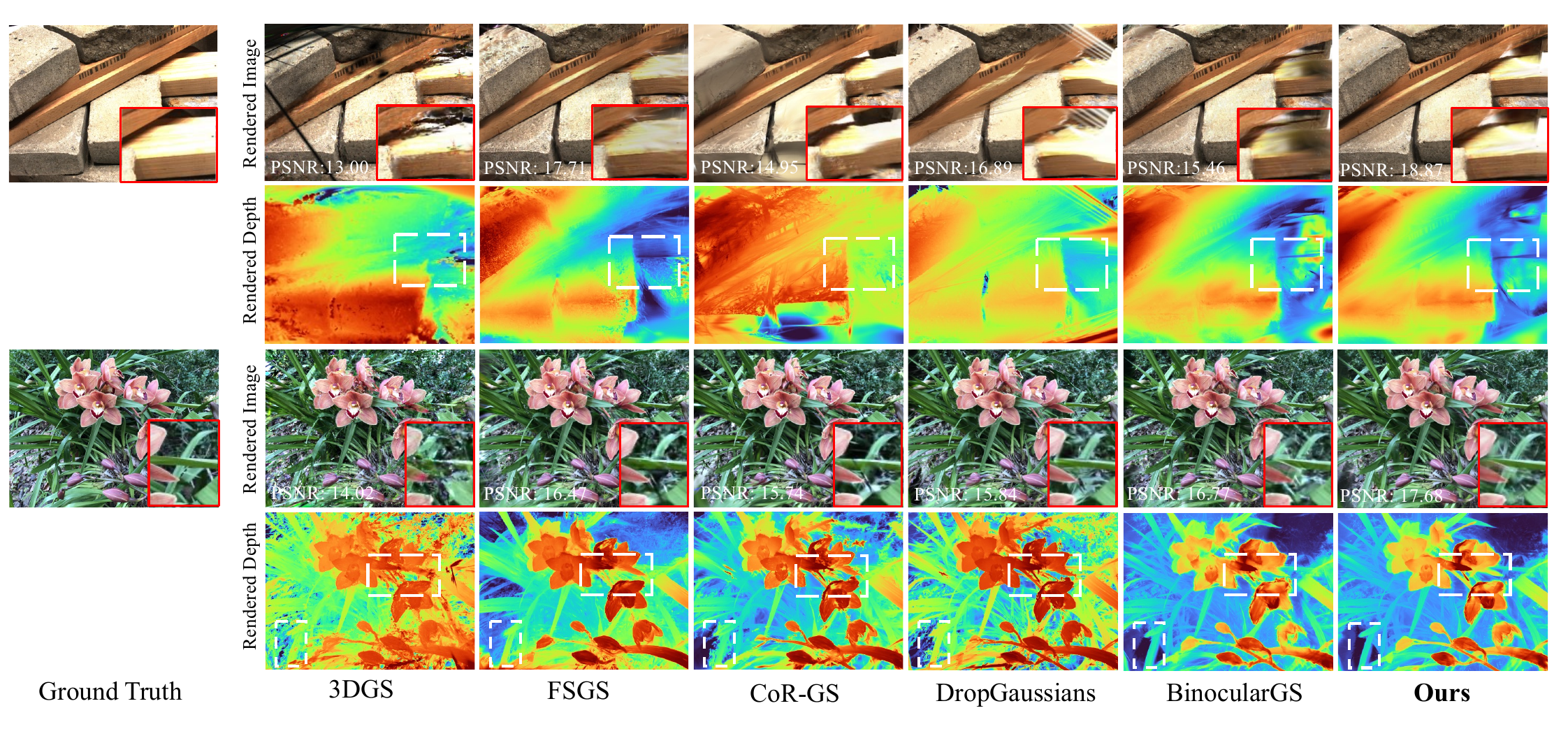}
}

\vspace{-0.3cm}
\centering
\captionsetup{hypcap=false}


\captionof{figure}{\textbf{Novel view synthesis quality under sparse inputs.} 
Our method produces superior photometric quality with faithful texture recovery 
in challenging regions (\eg, leaf gaps in zoomed insets and weakly-textured 
surfaces), enabled by more accurate 
geometry with sharper structural boundaries, outperforming state-of-the-art 
sparse-view 3DGS approaches~\cite{kerbl20233d,zhu2024fsgs,zhang2024cor,park2025dropgaussian,han2024binocular}.
}

\vspace{10pt}

\label{fig:head}
}

\makeatother

\maketitle
\input{secs/0_abstract}
\input{secs/1_intro}

\input{secs/2_related}
\input{secs/3_method}

\input{secs/4_exp}
\input{secs/5_conc}


{
    \small
    \bibliographystyle{ieeenat_fullname}
    \bibliography{main}
}


\end{document}

%% file: secs/0_abstract.tex
\begin{abstract}
3D Gaussian Splatting (3DGS) represents scenes through primitives with coupled intrinsic properties: geometric attributes (position, covariance, opacity) and appearance attributes (view-dependent color). 
Faithful reconstruction requires \textbf{intrinsic geometry-appearance consistency}, where geometry accurately captures 3D structure while appearance reflects photometry. 
However, sparse observations lead to appearance overfitting and underconstrained geometry, causing severe novel-view artifacts.
We present ICO-GS (Intrinsic Geometry-Appearance Consistency Optimization for 3DGS), a principled framework that enforces this consistency through tightly coupled geometric regularization and appearance learning. 
Our approach first regularizes geometry via feature-based multi-view photometric 
constraints by employing pixel-wise top-$k$ selection to handle occlusions and 
edge-aware smoothness to preserve sharp structures.
Then appearance is coupled with geometry through cycle-consistency depth filtering, which identifies reliable regions to synthesize virtual views that propagate geometric correctness into appearance optimization. 
Experiments on LLFF, DTU, and Blender show ICO-GS substantially improves geometry 
and photometry, consistently outperforming existing sparse-view baselines, particularly in challenging weakly-textured regions.
\end{abstract}

%% file: secs/1_intro.tex
\section{Introduction}
\label{sec:introduction}

Novel view synthesis (NVS) has witnessed remarkable progress with 3DGS~\cite{kerbl20233d}, which represents scenes as 
collections of anisotropic 3D Gaussians to achieve photorealistic rendering 
at real-time speeds. 
While 3DGS achieves real-time rendering and high visual quality on densely 
captured scenes, its performance degrades dramatically under 
sparse-view settings commonly encountered in practical scenarios.

The degradation stems from a fundamental issue in 3DGS optimization: 
\textbf{lack of intrinsic consistency between geometry and appearance}.
In 3DGS, each primitive is parameterized by coupled intrinsic properties, 
including geometric attributes (position $\boldsymbol{\mu}$, covariance 
$\boldsymbol{\Sigma}$, opacity $\alpha$) defining its spatial structure, 
and appearance attributes (view-dependent color $\mathbf{c}(\mathbf{d})$) 
determining its photometric contribution.
For faithful scene reconstruction, these properties must satisfy 
\emph{intrinsic consistency}: geometry should accurately capture the underlying 
3D structure, while appearance should coherently reflect surface photometry 
across viewpoints.
However, sparse-view 3DGS violates this consistency. 
The standard 3DGS optimization relies on per-view photometric supervision that 
independently minimizes rendering loss for each training view.
With limited observations, this independent supervision allows appearance to 
overfit individual views by compensating for geometric errors, while 3D geometry 
remains severely underconstrained due to lack of explicit multi-view regularization.
This leads to internally inconsistent 
Gaussians that produce plausible renderings on training views but severe 
artifacts (floaters or blurriness) on novel views.
These challenges are particularly pronounced in weakly-textured regions, 
where the absence of distinctive appearance cues further exacerbates geometric 
ambiguity.

Addressing this requires jointly solving two coupled challenges. 
\textbf{First}, how to effectively constrain geometry under sparse observations? 
Recent studies have employed pretrained depth estimation models~\cite{zhu2024fsgs,li2024dngaussian,xiong2023sparsegs, xu2025depthsplat} 
to regularize 3DGS geometry, but such approaches suffer from scale ambiguity and 
noise introduced by imperfect pretrained models. 
Other methods~\cite{zhu2024fsgs,zheng2025nexusgs} rely on dense initialization, 
yet the initial geometric cues tend to be gradually forgotten during subsequent optimization.
\textbf{Second}, how to couple accurate geometry with reliable appearance optimization to prevent overfitting? 
BinocularGS~\cite{han2024binocular} builds virtual binocular pairs from rendered depth 
to enforce disparity consistency, but this approach relies on rendered depth whose 
reliability is not guaranteed, potentially propagating depth errors into appearance optimization 
and limiting rendering quality.

To this end, we propose \textbf{ICO-GS} (\textbf{I}ntrinsic Geometry-Appearance 
\textbf{C}onsistency \textbf{O}ptimization for Sparse-view \textbf{G}aussian 
\textbf{S}platting), a principled framework that restores intrinsic consistency 
in sparse-view 3DGS through synergistic geometry-appearance optimization.
Our key insight is that faithful geometry and appearance emerge from their 
\emph{mutual reinforcement}: well-constrained geometry guides appearance to 
learn view-consistent photometry, while reliable appearance supervision in 
turn refines geometry.
We realize this synergy through two coupled components.

\textbf{Geometric regularization via multi-view photometric consistency.}
Under ideal conditions, a 3D point observed from multiple viewpoints should 
exhibit photometric consistency. However, illumination variations, occlusions, 
and monocularly visible regions violate this assumption. 
We therefore adopt feature-based multi-view consistency to regularize geometry, 
mitigating the impact of lighting and imaging variations. 
To handle occlusions prevalent in sparse-view settings, we employ pixel-wise 
top-$k$ selection: for each pixel, we compute photometric errors across all 
source views and retain only the $k$ most consistent ones, robustly filtering 
occluded or unreliable observations. 
For monocularly visible regions, we further incorporate an edge-aware depth 
smoothness term that enforces local coherence while preserving sharp geometric 
boundaries. 
These complementary constraints yield geometry that is both multi-view consistent 
and structurally sound, proving particularly effective in weakly-textured 
regions where photometric cues are scarce.

\textbf{Geometry-guided appearance optimization via virtual view consistency.}
To couple geometry with appearance, we leverage the regularized depth to 
synthesize virtual views, propagating geometric constraints to appearance.
However, naively using all depth estimates introduces noise from uncertain regions.
We address this via cycle-consistency depth filtering: we project each pixel's 
depth to source views and back-project to the original view, retaining only 
pixels with consistent spatial locations.
The filtered depth then guides virtual view synthesis through depth-conditioned 
warping, providing geometric supervision that encourages appearance to capture 
view-consistent photometry rather than overfit individual observations.

Our contributions are summarized as follows:
\begin{itemize}[leftmargin=*,nosep]
      \item We identify \emph{intrinsic consistency}, defined as the coupled correctness 
      of geometry and appearance, as a fundamental principle for sparse-view 3DGS, 
      and reveal how its violation causes severe degradation in novel views.
    \item We address geometry underconstraint through feature-based multi-view 
          photometric regularization with pixel-wise top-$k$ consistency and 
          edge-preserving smoothness.
    \item We prevent appearance overfitting through geometry-guided optimization 
          that synthesizes virtual views from cycle-filtered depth, coupling 
          geometric accuracy with photometric quality.
    \item Extensive experiments validate that ICO-GS achieves state-of-the-art sparse-view novel view synthesis across diverse scenes, particularly on weakly-textured data, with 1.1 dB PSNR gains over prior arts on 3-view DTU scenarios.
\end{itemize}

%% file: secs/2_related.tex
\section{Related Work}
\label{sec:related}
\subsection{Novel View Synthesis}

\begin{figure*}[t]
	\vspace{-0.3cm}
	\begin{center}
	   \includegraphics[trim={0cm 0cm 0cm 0cm},clip,width=1.0\linewidth]{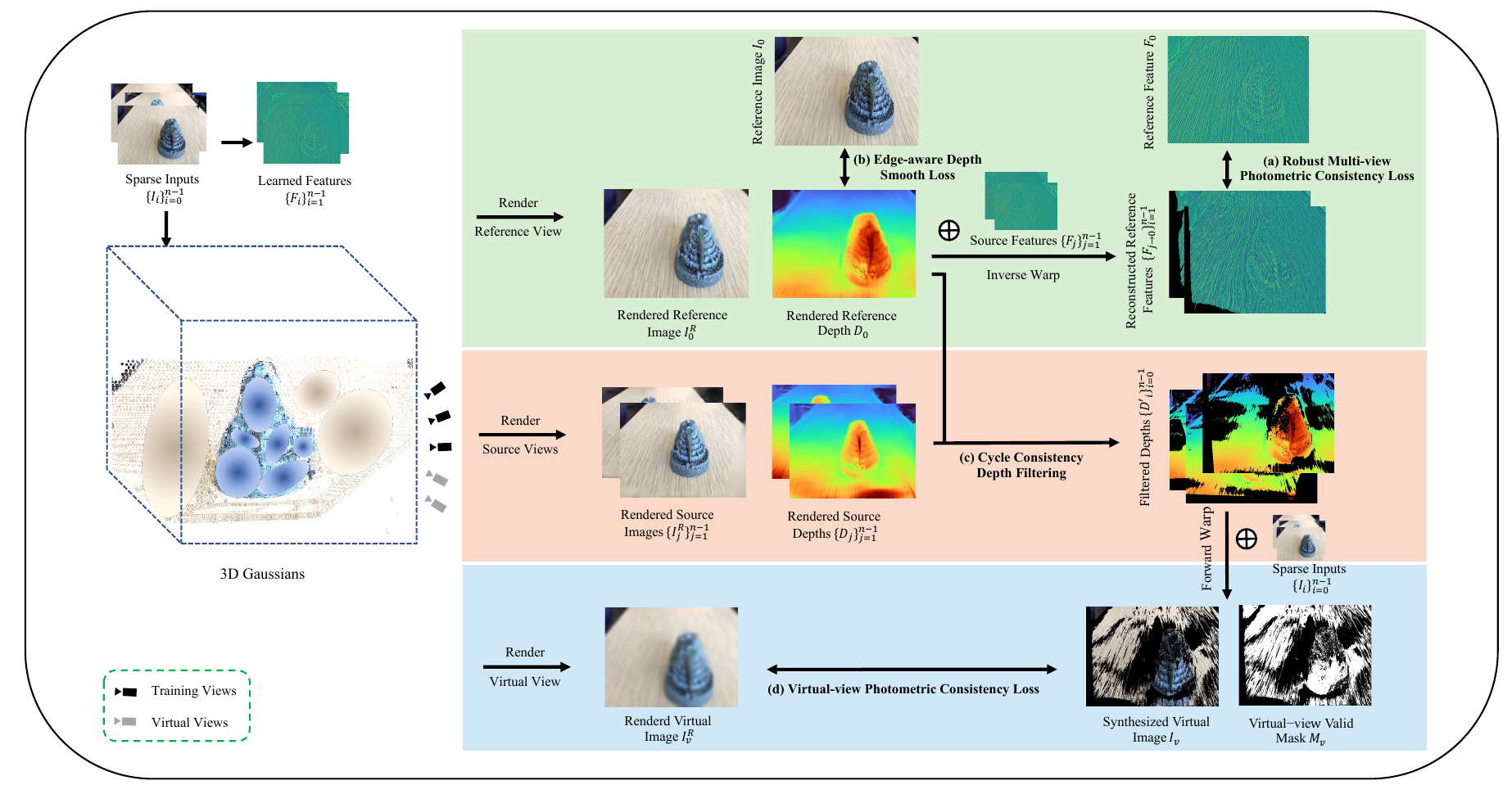}
	\end{center}
	\vspace{-0.4cm}
	\caption{\textbf{Framework of ICO-GS.} 
	Given sparse input views, we initialize 3D Gaussians and extract deep features. 
	Our method enforces intrinsic geometry-appearance consistency through two synergistic 
	components: 
	\textbf{(1) Robust Geometric Regularization.} Source features are warped to reference 
	views via rendered depth, establishing occlusion-aware multi-view constraints through: 
	(a) \textit{Robust Multi-view Photometric Consistency} that employs pixel-wise top-$k$ 
	selection for occlusion handling, and (b) \textit{Edge-aware Depth Smoothness} that 
	preserves sharp geometric structures. 
	\textbf{(2) Geometry-Guided Appearance Optimization.} We leverage geometrically reliable 
	regions identified by (c) \textit{Cycle Consistency Depth Filtering} to synthesize 
	virtual views, then apply (d) \textit{Virtual-view Photometric Consistency} between 
	synthesized and rendered images to propagate geometric correctness into appearance 
	learning.
	}
	
	\label{fig:framework}
	\vspace{-0.4cm}
  \end{figure*}

Novel View Synthesis (NVS) seeks to generate novel perspectives of a scene given a collection of input images. The introduction of Neural Radiance Fields (NeRF)~\cite{mildenhall2021nerf} revolutionized this domain by representing scenes as volumetric radiance fields parameterized by multilayer perceptrons. Following this seminal work, substantial research efforts have been devoted to enhancing various aspects of NeRF, including photorealistic quality~\cite{barron2021mip,barron2022mip360,barron2023zip}, rendering efficiency~\cite{garbin2021fastnerf,fridovich2022plenoxels,muller2022instant,chen2022tensorf}, and generalization to complex scenarios such as dynamic environments~\cite{park2021nerfies,pumarola2021d} and unconstrained captures~\cite{martin2021nerf}. Despite achieving impressive visual fidelity, NeRF-based approaches inherently suffer from prohibitive computational demands.To overcome these computational bottlenecks, 3D Gaussian Splatting (3DGS)~\cite{kerbl20233d} emerged as a promising alternative by representing scenes with explicit 3D Gaussian primitives rather than implicit neural functions, enabling real-time novel view synthesis with comparable quality. This breakthrough has inspired numerous follow-up works that extend 3DGS to various challenging scenarios~\cite{yu2024mip, lu2024scaffold, matsuki2024gaussian, kocabas2024hugs, fu2024colmap, yan2024street}, including anti-aliasing~\cite{yu2024mip}, surface reconstruction~\cite{yu2024gaussian, huang20242d, li2024monogsdf}, dynamic scene modeling~\cite{wu20244d}, and memory-efficient representations~\cite{chen2024hac}. 
Despite achieving impressive results on densely captured scenes, both NeRF and 
3DGS exhibit severe degradation under sparse views, as their standard 
optimization relies on dense multi-view supervision to disambiguate geometry 
and appearance.

\subsection{Sparse-view Novel View Synthesis}
Neural rendering methods like NeRF and 3DGS optimize appearance independently per view through photometric losses, allowing appearance to overfit training observations while geometry remains  underconstrained, violating the multi-view consistency principle essential for sparse-view novel view synthesis.
Early sparse-view NeRF methods introduced various regularization strategies, including semantic consistency via CLIP embeddings~\cite{jain2021dietnerf}, patch-based geometric regularization~\cite{niemeyer2022regnerf}, and frequency regularization~\cite{yang2023freenerf} to constrain this ill-posed optimization. 

For 3DGS, several works analyzed the overfitting issue in sparse-view 3DGS and include dual-model regularization~\cite{zhang2024cor} or dropout~\cite{park2025dropgaussian,xu2025dropoutgs,chen2025quantifying} to mitigate it. 
However, these methods lack analysis of the intrinsic geometry of the Gaussian primitives.
Alternative approaches~\cite{zhu2024fsgs,li2024dngaussian,xiong2023sparsegs, paliwal2024coherentgs, chung2024depth} integrate external monocular depth priors from pre-trained estimators~\cite{ranftl2021vision, cao2022mvsformer}, such priors are subject to scale ambiguity and can introduce erroneous or misleading noise. 
Some methods propose dense initialization~\cite{zheng2025nexusgs} and binocular warping consistency~\cite{han2024binocular}, but they cannot guarantee that the geometry of the Gaussian primitives remains accurate throughout iterative optimization, which limits their effectiveness.
Unlike prior methods that do not adequately enforce intrinsic consistency between 
accurate geometry and reliable appearance, we propose a framework grounded in 
intrinsic consistency optimization, based on the principle that geometry and 
appearance should be mutually correct and reinforcing.

%% file: secs/3_method.tex
\section{Method}
\label{sec:method}
In this section, we introduce \textbf{ICO-GS}, a sparse-view reconstruction approach 
that enforces intrinsic consistency by coupling geometric regularization with 
appearance optimization. 
The overall framework is illustrated in \cref{fig:framework}. 
We first analyze the geometry-appearance discrepancy phenomenon under sparse-view 
settings (\cref{subsec:motivation}), then elaborate our robust geometric 
regularization (\cref{subsec:geometry}) and geometry-guided appearance 
optimization (\cref{subsec:appearance}). 
The complete optimization pipeline is described in \cref{subsec:optimization}.

\subsection{Motivation}
\label{subsec:motivation}

\begin{figure}[t]
  \vspace{-0.3cm}
  \begin{center}
     \includegraphics[trim={0cm 0cm 0cm 0cm},clip,width=1.0\linewidth]{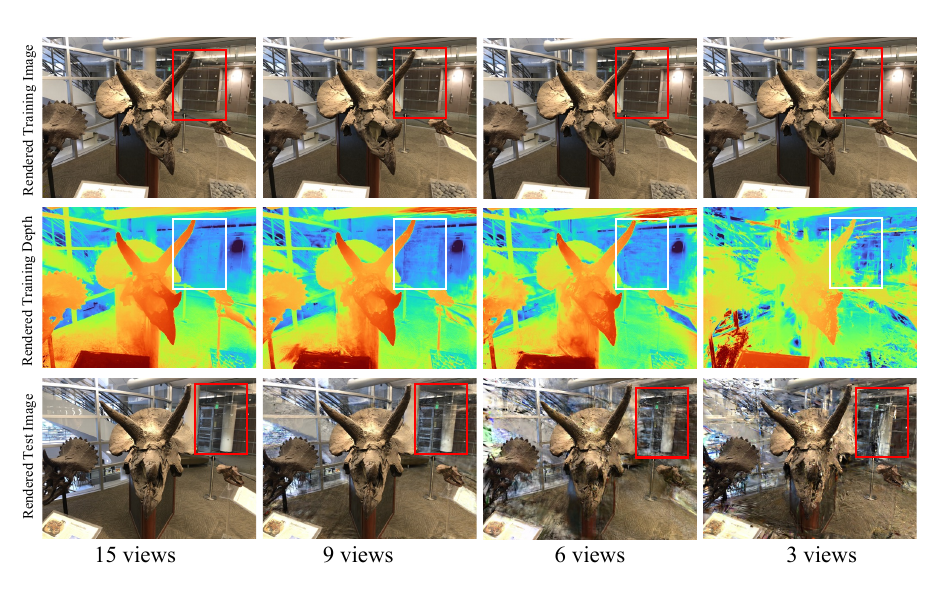}
  \end{center}
  \vspace{-0.8cm}
  \caption{\textbf{Geometry-appearance discrepancy under sparse-view settings.}
  From top to bottom: RGB on training views, depth on training views, and RGB on test 
  views, rendered by 3D Gaussian Splatting~\cite{kerbl20233d} with varying training 
  view densities. With decreasing views, training-view appearance (top) remains 
  well-fitted, but depth quality (middle) collapses with noise and floaters
  due to insufficient multi-view constraints. This geometry-appearance discrepancy 
  leads to severe artifacts in novel-view rendering (bottom).
}
  \label{fig:motivation}
  \vspace{-0.6cm}
  \end{figure}

  Despite success in dense-view reconstruction, 3DGS~\cite{kerbl20233d} severely 
  overfits under sparse observations. 
  We reveal a critical \textbf{geometry-appearance discrepancy}: as view count 
  decreases, appearance quality remains superficially high on training views while 
  geometric accuracy collapses, causing severe artifacts on novel views 
  (\cref{fig:motivation}).
  
  We identify two underlying deficiencies:
  
  \noindent\textbf{Insufficient Geometric Constraints.} 
  In dense-view settings, each 3D point is observed by multiple cameras (typically 10+), 
  providing strong multi-view constraints on Gaussian positions. 
  However, with sparse views, each point is visible in only 2--3 views, leading to 
  severe geometric ambiguity. 
  Mathematically, the photometric loss in \cite{kerbl20233d} only constrains the 
  \emph{projected} appearance of Gaussians, not their depth. 
  Consequently, Gaussians can be placed at \emph{any} position along the camera ray 
  while maintaining zero photometric error. 
  This depth ambiguity explains the noisy geometry in \cref{fig:motivation}: without 
  sufficient multi-view overlap, optimization lacks constraints to determine correct 
  3D positions.
    
  \noindent\textbf{Unreliable Appearance-Geometry Coupling.} 
  Moreover, the 3DGS representation does not inherently ensure geometry-appearance 
  consistency. 
  In principle, when a Gaussian is misplaced, the photometric loss should drive the 
  optimizer to correct its position. 
  However, we observe a problematic shortcut: instead of moving the Gaussian, the 
  optimizer adjusts its color and opacity to compensate for geometric errors, which 
  we term \emph{appearance compensation}.
  As shown in \cref{fig:motivation}, training views achieve high PSNR despite 
  severely incorrect geometry (noisy depth), demonstrating that appearance parameters 
  mask geometric mistakes. 
  This becomes particularly severe under sparse views, where weak geometric constraints 
  enable the model to overfit through appearance manipulation rather than learning 
  correct 3D structure.
  
  To address these deficiencies, we propose \textbf{ICO-GS}, which restores intrinsic 
  consistency by coupling robust geometric regularization (\cref{subsec:geometry}) 
  with geometry-guided appearance optimization (\cref{subsec:appearance}). 
  These components work synergistically to promote both geometry and appearance.

\subsection{Robust Geometric Regularization}
\label{subsec:geometry}
As identified in \cref{subsec:motivation}, sparse-view 3DGS suffers from insufficient geometric constraints. 
BinocularGS~\cite{han2024binocular} attempts to address this by enforcing stereo consistency via depth-warped virtual views. 
However, this approach suffers from a fundamental circular dependency: unreliable depth produces misaligned virtual views, which in turn provide corrupted supervision that further degrades geometry.

We break this loop via \emph{robust geometry regularization}, which warps pixels between training views according to rendered depth and penalizes inconsistencies. 
To handle illumination variations and occlusions that undermine naïve photometric matching, we introduce robust multi-view photometric consistency (\cref{subsubsec:mpc}) enhanced with edge-aware depth smoothness (\cref{subsubsec:smooth}).

\subsubsection{Multi-view Photometric Consistency}
\label{subsubsec:mpc}

Given $n$ sparse training views $\{I_i\}_{i=0}^{n-1}$, our goal is to regularize the rendered Gaussian depths $\{D_i\}_{i=0}^{n-1}$ via multi-view photometric consistency.
Take one reference view $I_{\text{0}}$ and its corresponding source views $\{I_j\}_{j=1}^{n-1}$ for example, we first render the depth map $D_{\text{0}}$ via alpha-blending rendering. 
Then for each pixel $p$ in the reference image $I_{\text{0}}$, its corresponding pixel $p'_{j}$ in the source images $\{I_j\}_{j=1}^{n-1}$ can be computed via:
\begin{equation}
  \vspace{-0.2cm}
  \label{eq:inverse_warp} 
    p'_{j} = KT_{0 \rightarrow j}(D_{0}(p) \cdot K^{-1} p),    
\end{equation}
where $K, T$ denote the associated intrinsic and the relative transformation.
We enforce inverse warp from the source views to reference views to acquire the reconstructed reference images
$\{I_{j \rightarrow 0} \}_{j=1}^{n-1}$ with a binary validity mask $\{M_j\}_{j=1}^{n-1}$ indicating valid  projected pixels during warping. 
For ideal unoccluded Lambertian surfaces, pixels  in $\{ I_0, \{I_{j \rightarrow 0} \}_{j=1}^{n-1}  \}$ should be photometrically consistent. 
The multi-view photometric loss enforces this:
\begin{equation}
  \vspace{-0.2cm}
\label{eq:mpc_rgb}
L = \frac{1}{n-1} \sum_{j=1}^{n-1}  \frac  {\left\| (I_{j \rightarrow 0} - I_{\text{0}}) \odot M_j  \right\|_1} {\left\|  M_j \right\|_1}.
\end{equation}

\noindent\textbf{Illumination-robust Feature Matching.}
Equation~\ref{eq:mpc_rgb} relies on RGB consistency, which is fragile to 
lighting variations, shadows, and specular reflections common in real scenes. 
To achieve robust geometric supervision, we replace it with feature-based matching using a frozen pre-trained feature network in~\cite{gu2020cascade}:
\begin{equation}
\label{eq:mpc_feat}
\vspace{-0.2cm}
L = \frac{1}{n-1} \sum_{j=1}^{n-1}  \frac  {\left\| \frac{1}{2}\left(1 - \cos(\mathcal{F}_{\text{0}}, \mathcal{F}_{j \rightarrow 0 } ) \right) \odot M_j  \right\|_1} {\left\|  M_j \right\|_1},
\end{equation}
where $\mathcal{F}_{\text{0}}$ and ${ \mathcal{F}_{j \rightarrow 0 } }$ are features extracted from the reference view and features warped from source views $j$ to the reference view $0$. 
Since features are computed once during preprocessing and remain frozen during training, this incurs negligible computational overhead while significantly improving robustness to illumination changes.

\noindent\textbf{Occlusion-aware Photometric Consistency.}
Occlusions cause photometric consistency to \textbf{fail} even with accurate depth. 
We address this via \emph{pixel-wise top-k selection}, which adaptively chooses the most reliable correspondences from visible source views.
For each reference pixel $p$, we identify the top-$k$ most consistent correspondences across all warped source features $\{  \mathcal{F}_{j \rightarrow 0 }  \}_{j=1}^{n-1}$:
\begin{equation}
  \begin{aligned}
  \mathcal{T}_k(p) = \underset{\substack{S \subset \{1,\ldots,n-1\} \\ |S|=k}}{\arg\min} 
      \sum_{j \in S} \left\| \frac{1}{2}\left(1 - \cos(\mathcal{F}_{\text{0}}(p), \mathcal{F}_{j \rightarrow 0}(p))  \right) \right\|_1.
  \end{aligned}
  \end{equation}
This reformulates \cref{eq:mpc_feat} as an adaptive aggregation over $\mathcal{T}_k(p)$:
\begin{equation}
\label{eq:mpc_final}
\mathcal{L}_{\text{mpc}}^{\text{Fea}}(p) = \frac{1}{k} \sum_{j \in \mathcal{T}_k(p)} \left\| \frac{1}{2} \left(1 - \cos(\mathcal{F}_{\text{0}}(p), \mathcal{F}_{j \rightarrow 0}(p)) \right) \right\|_1.
\end{equation}
This pixel-wise selection naturally handles spatially-varying occlusions: for pixels occluded in half the views, the remaining visible views still provide valid supervision. 
Setting $k = \lceil (n-1)/2 \rceil$ balances coverage and outlier rejection.

\subsubsection{Edge-aware Depth Smoothness.}
\label{subsubsec:smooth}

In regions visible from only one views, multi-view photometric consistency fails to provide sufficient geometric constraints. 
We therefore regularize these under-constrained areas with edge-aware depth smoothness:
\begin{equation}
  \vspace{-0.2cm}
\label{eq:smooth}
\mathcal{L}_{\text{smooth}} = \sum_{p} \left\| \nabla D_{\text{0}}(p) \right\|_1 \cdot \exp\left( -\alpha \left\| \nabla I_{\text{0}}(p) \right\|_1 \right),
\end{equation}
where $\nabla D$ and $\nabla I$ denote the depth and image gradients, respectively, and $\alpha=1$ controls edge sensitivity. This encourages smooth depth in textureless regions while preserving discontinuities at object boundaries.

By enforcing geometric consistency through this robust regularization, our 
method significantly improves novel view synthesis quality in weakly-textured regions—
areas where 3D Gaussian Splatting typically struggles due to insufficient photometric 
constraints.
The geometric regularization provides reliable supervision even when 
appearance cues are ambiguous.

\subsection{Geometry-guided Appearance Optimization }
\label{subsec:appearance}
We propose to leverage regularized geometry for appearance optimization through 
virtual-view sampling. 
Existing methods~\cite{zhu2024fsgs,li2024dngaussian,xu2024mvpgs,han2024binocular} 
render virtual novel views to mitigate texture under-constraint in sparse settings, 
enforcing regularization via monocular or binocular depth consistency.
Yet they 
are limited by: scale ambiguity in monocular depth~\cite{zhu2024fsgs,li2024dngaussian}, 
noise in MVS priors~\cite{xu2024mvpgs}, and restricted diversity with inaccurate 
rendered depth~\cite{han2024binocular}.

We instead utilize our regularized Gaussian geometry to enable flexible, reliable 
virtual-view supervision. Our approach comprises: cycle-consistency 
filtering to identify valid depth regions (\cref{sec:depth_filter}), and 
appearance supervision via virtual-view photometric consistency over these 
validated regions (\cref{sec:reliable_synthesis}).

\subsubsection{Cycle Consistency Depth Filtering}
\label{sec:depth_filter}
To ensure geometry reliability, we validate rendered depth through cycle-consistency filtering before synthesizing virtual views.
Given a reference view $I_{\text{0}}$, source views $\{I_j\}_{j=1}^{n-1}$, camera 
intrinsic $K$, relative transformations $\{T_{0 \rightarrow j} \}_{j=1}^{n-1}$, and rendered depth 
maps $\{D_i\}_{i=0}^{n-1}$ from Gaussian splatting, we perform forward-backward 
warping.
For each pixel $p$ in $I_{\text{0}}$, we first forward warp to source 
view $I_j$ using depth $D_0(p)$ to obtain projected pixel $p'_{j}$ (see \cref{eq:inverse_warp}). 
We then backward warp $p'_{j}$ to $I_{\text{0}}$ using source depth $D_j(p'_{j})$ 
to obtain reprojected pixel $p''_{j}$ and depth $\tilde{D}_j(p)$:
\begin{equation}
  \vspace{-0.2cm}
\label{eq:inverse_warp_back}
p''_j = KT^{-1}_{0 \rightarrow j}D_j(p'_{j})K^{-1}p'_{j}.
\end{equation}
The depth error between original and reprojected depth measures geometric consistency:
\begin{equation}
\label{eq:depth_error}
\vspace{-0.2cm}
e_j(p) = \left| D_{\text{0}}(p) - \tilde{D}_j(p) \right|.
\end{equation}
A pixel is reliable if its depth error falls below threshold $\tau_d$ for at 
least $m$ of the $n-1$ source views:
\begin{equation}
  \vspace{-0.2cm}
\label{eq:depth_mask}
\mathcal{M}_{\text{reliable}}(p) = \mathbb{I}\left[ \sum_{j=1}^{n-1} \mathbb{I}[e_j(p) < \tau_d] \geq m \right],
\end{equation}
where $m = \lceil (n-1)/2 \rceil$ ensures consistency with at least half the 
sources, and $\tau_d = 0.01 \cdot \max(D_{\text{0}})$. This binary mask 
$\mathcal{M}_{\text{reliable}}$ identifies regions where rendered depth $D_0$ is 
validated by cycle consistency, ensuring subsequent warping produces views 
aligned with true scene structure.

\subsubsection{Virtual-view Photometric Consistency}
\label{sec:reliable_synthesis}
With reliable depth identified, we propagate accurate geometry to unseen appearances via virtual-view photometric consistency. 
Unlike prior stereo-pair approaches~\cite{han2024binocular}, we sample virtual poses $\{\mathcal{P}_v\}_{v=1}^{N_v}$ across a wider range: for each reference position $\mathbf{x}$, we randomly sample within a sphere of radius $r$, providing sufficient viewpoint diversity.

For each virtual view, we forward warp (\cref{eq:inverse_warp}) pixels from all training images $\{I_i\}_{i=0}^{n-1}$ using masked depths $\{\mathcal{M}_{\text{reliable}}^{i} \odot D_i\}_{i=0}^{n-1}$ to synthesize virtual image $\mathcal{I}_{v}$ with validity mask $M_v$, excluding unreliable regions to prevent geometric errors from contaminating supervision.

\begin{table}[t]
  \centering
  \caption{\textbf{Quantitative comparisons on LLFF~\cite{mildenhall2019local} dataset under sparse view settings. }}
  \label{tab:llff_results}
  \resizebox{1.0\linewidth}{!}{
  \begin{tabular}{l|ccc|ccc|ccc}
  \toprule
  \multirow{2}{*}{Methods} & \multicolumn{3}{c}{PSNR↑} & \multicolumn{3}{c}{SSIM↑} & \multicolumn{3}{c}{LPIPS↓} \\

  \cmidrule{2-10}
  & 3-view & 6-view & 9-view & 3-view & 6-view & 9-view & 3-view & 6-view & 9-view \\
  \midrule
  DietNeRF~\cite{jain2021dietnerf} & 14.94 & 21.75 & 24.28 & 0.370 & 0.717 & 0.801 & 0.496 & 0.248 & 0.183 \\
  RegNeRF~\cite{niemeyer2022regnerf} & 19.08 & 23.10 & 24.86 & 0.587 & 0.760 & 0.820 & 0.336 & 0.206 & 0.161\\
  FreeNeRF~\cite{yang2023freenerf} & 19.63 & 23.73 & 25.13 & 0.612 & 0.779 & 0.827 & 0.308 & 0.195 & 0.160 \\
  SparseNeRF~\cite{wang2023sparsenerf} & 19.86 & 23.26 & 24.27 & 0.714 & 0.741 & 0.781 & 0.243 & 0.235 & 0.228 \\
  \midrule
  3DGS~\cite{kerbl20233d} & 15.52 & 19.45 & 21.13 & 0.405 & 0.627 & 0.715 & 0.408 & 0.268 & 0.214 \\
  FSGS~\cite{zhu2024fsgs} & 20.31 & 24.20 & 25.32 & 0.652 & 0.811 & 0.856 & 0.288 & 0.173 & 0.136 \\
  DNGaussian~\cite{li2024dngaussian} & 19.12 & 22.18 & 23.17 & 0.591 & 0.755 & 0.788 & 0.294 & 0.198 & 0.180\\
  CoR-GS~\cite{zhang2024cor} & 20.45 & 24.49 & 26.06 & 0.712 & 0.837 & 0.874 & 0.196 & 0.115 & \colorbox{thirdplace}{0.089} \\
  BinocularGS~\cite{han2024binocular} & \colorbox{secondplace}{21.44} & \colorbox{thirdplace}{24.87} & 26.17 & \colorbox{secondplace}{0.751} & \colorbox{thirdplace}{0.845} & \colorbox{thirdplace}{0.877} & \colorbox{secondplace}{0.168} & \colorbox{firstplace}{0.106} & 0.090 \\
  DropGaussians~\cite{park2025dropgaussian} & 20.76 & 24.74 & \colorbox{thirdplace}{26.21} & 0.713 & 0.837 & 0.874 & 0.200 & 0.117 & \colorbox{secondplace}{0.088} \\
  NexusGS~\cite{zheng2025nexusgs} & 21.07 & - & - & 0.738 & - & - & \colorbox{thirdplace}{0.177} & - & - \\
  ComapGS~\cite{jang2025comapgs} & \colorbox{thirdplace}{21.11} & \colorbox{secondplace}{25.20} & \colorbox{firstplace}{26.73} & \colorbox{thirdplace}{0.747} & \colorbox{secondplace}{0.854} & \colorbox{firstplace}{0.886} & 0.182 & \colorbox{secondplace}{0.108} & \colorbox{firstplace}{0.082} \\
  \midrule
  \textbf{Ours} & \colorbox{firstplace}{\textbf{22.20}} & \colorbox{firstplace}{\textbf{25.37}} & \colorbox{secondplace}{\textbf{26.45}} & \colorbox{firstplace}{\textbf{0.778}} & \colorbox{firstplace}{\textbf{0.856}} & \colorbox{secondplace}{\textbf{0.881}} & \colorbox{firstplace}{\textbf{0.157}} & \colorbox{thirdplace}{\textbf{0.109}} & {\textbf{0.096}} \\
  \bottomrule
  \end{tabular}
  }
  \end{table}

\begin{figure*}[t]
\vspace{-0.2cm}
\begin{center}
   \includegraphics[trim={2cm 0cm 2cm 0cm},clip,width=0.72\linewidth]{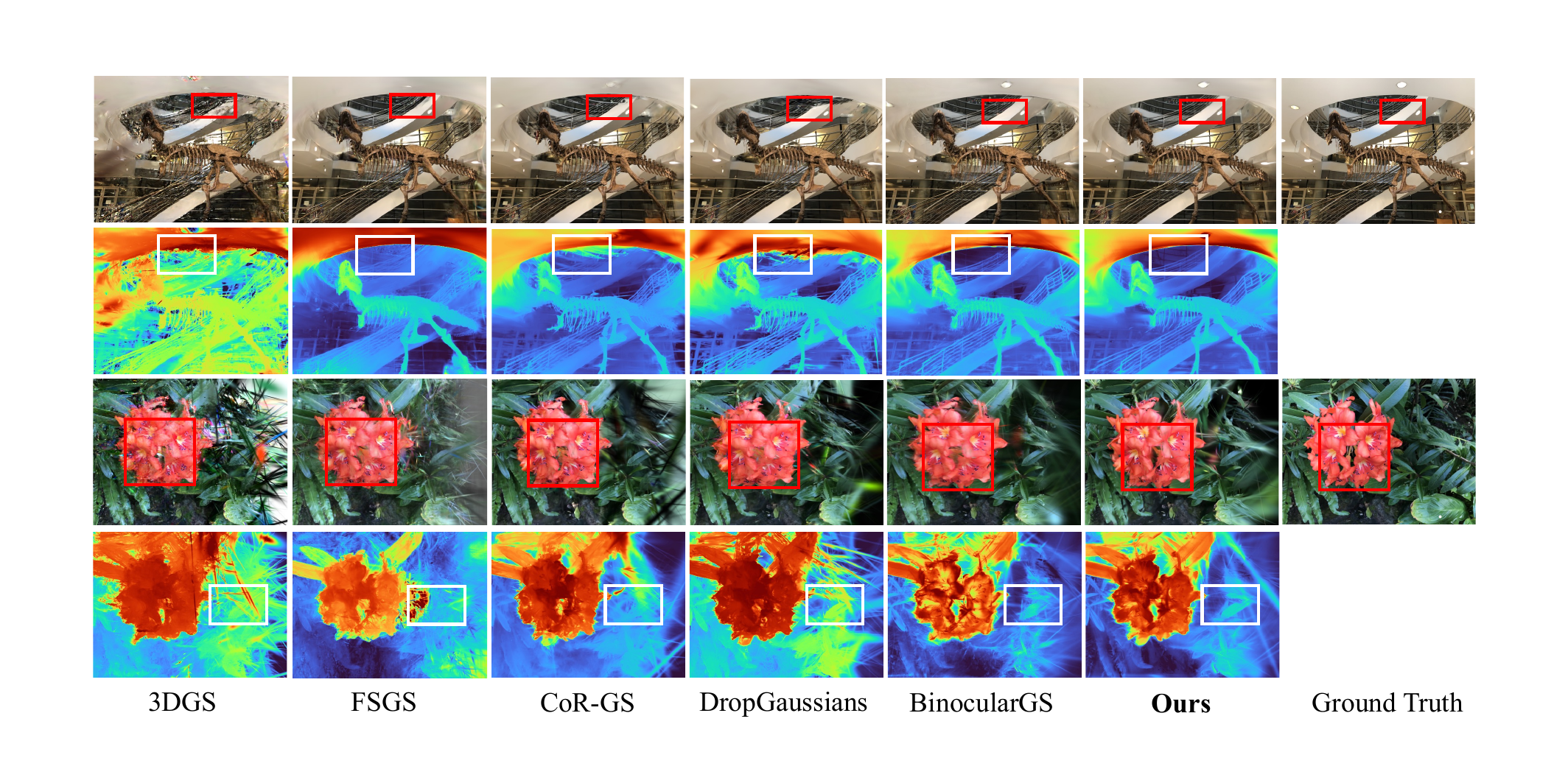}
\end{center}
\vspace{-1cm}
\caption{\textbf{Visual comparison on LLFF~\cite{mildenhall2019local} dataset.}}
\label{fig:llffcom}
\vspace{-0.3cm}
\end{figure*}

\noindent\textbf{Vitural-view Photometric Consistency Loss.}
The synthesized virtual images are incorporated into training for appearance optimization. 
For each virtual view, we render the Gaussian to acquire the rendered virtual image $\mathcal{I}_{v}^{R}$ from the 
virtual pose $\mathcal{P}_v$ and enforce photometric consistency on valid pixels:
\begin{equation}
  \vspace{-0.2cm}
\label{eq:aug_loss}
L_{app} = \sum_{p \in \mathcal{M}_{\text{v}}}   \left\| \mathcal{I}_{v}(p) - \mathcal{I}_{v}^{R}(p) \right\|_1.
\end{equation}
This serves two purposes: it provides additional observations to optimize appearance in unseen views and prevent overfitting, and conversely, it constrains geometry through supervision from novel viewpoints. 
Critically, because virtual-view images are synthesized from reliability-filtered depth, they provide clean supervision without the geometric distortions introduced by prior methods relying on unreliable depth predictions.

\subsection{Overall Pipeline}
\label{subsec:optimization}
We integrate geometric regularization and geometry-guided appearance 
optimization via curriculum learning~\cite{bengio2009curriculum}.

\noindent\textbf{Training Objective.} The complete loss combines four terms:
  \begin{equation}
    \begin{split}
    \mathcal{L}_{\text{total}} = 
    \mathcal{L}_{\text{3DGS}} 
    &+ \mathcal{L}_{\text{consis}} 
    + \lambda_{\text{mpc}} \mathcal{L}_{\text{mpc}}^{\text{Fea}} \\
    &+ \lambda_{\text{smooth}} \mathcal{L}_{\text{smooth}} 
    + \lambda_{\text{app}} \mathcal{L}_{\text{app}},
    \end{split}
    \end{equation} 
    where $\mathcal{L}_{\text{3DGS}}$ is the base photometric loss, 
    $\mathcal{L}_{\text{consis}}$ enforces binocular consistency inherited from the baseline~\cite{han2024binocular}, 
    $\mathcal{L}_{\text{mpc}}^{\text{Fea}} + \lambda_{\text{smooth}}\mathcal{L}_{\text{smooth}}$ enforces geometry regularization (\cref{subsec:geometry}), 
    and $\mathcal{L}_{\text{app}}$ optimizes appearance in unseen views (\cref{subsec:appearance}).

\noindent\textbf{Optimization.}
We employ a three-stage curriculum: Stage 1 optimizes $\mathcal{L}_{\text{3DGS}}$ to establish coarse geometry; Stage 2  activates geometric regularization $\lambda_{\text{mpc}}  \mathcal{L}_{\text{mpc}}^{\text{Fea}} + \lambda_{\text{smooth}}\mathcal{L}_{\text{smooth}}$; 
Stage 3 adds appearance supervision $\lambda_{\text{app}} \mathcal{L}_{\text{app}}$ from virtual views. 
This staged approach ensures stable convergence.

%% file: secs/4_exp.tex
\section{Experiments}
\label{sec:exp}

\begin{table}[t]
    \centering
    \caption{\textbf{Quantitative comparisons on DTU dataset under sparse view settings. }}
    \label{tab:dtu_results}
    \resizebox{1.0\linewidth}{!}{
    \begin{tabular}{l|ccc|ccc|ccc}
    \toprule
    \multirow{2}{*}{Methods} & \multicolumn{3}{c}{PSNR↑} & \multicolumn{3}{c}{SSIM↑} & \multicolumn{3}{c}{LPIPS↓} \\
  
    \cmidrule{2-10}
    & 3-view & 6-view & 9-view & 3-view & 6-view & 9-view & 3-view & 6-view & 9-view \\
    \midrule
    DietNeRF~\cite{jain2021dietnerf} & 11.85 & 20.63 & 23.83 & 0.633 & 0.778 & 0.823 & 0.214 & 0.201 & 0.173 \\
    RegNeRF~\cite{niemeyer2022regnerf} & 18.89 & 22.20 & 24.93 & 0.745 & 0.841 & 0.884 & 0.190 & 0.117 & \colorbox{thirdplace}{0.089} \\
    FreeNeRF~\cite{yang2023freenerf} & 19.52 & 23.25 & 25.38 & 0.787 & 0.844 & 0.888 & 0.173 & 0.131 & 0.102 \\
    SparseNeRF~\cite{wang2023sparsenerf} & 19.47 & - & - & 0.829 & - & - & 0.183 & - & - \\
    \midrule
    3DGS~\cite{kerbl20233d} & 10.99 & 20.33 & 22.90 & 0.585 & 0.776 & 0.816 & 0.313 & 0.223& 0.173 \\
    FSGS~\cite{zhu2024fsgs} & 17.34 & 21.55 & 24.33 & {0.818} & \colorbox{thirdplace}{0.880} & \colorbox{thirdplace}{0.911} & 0.169 & 0.127 & 0.106 \\
    DNGaussian~\cite{li2024dngaussian} & 18.91 & 22.10 & 23.94 & 0.790 & 0.851 & 0.887 & 0.176 & 0.148 & 0.131 \\

    CoR-GS~\cite{zhang2024cor} & {19.21} & \colorbox{secondplace}{24.51} & \colorbox{secondplace}{27.18} & 0.853 & \colorbox{secondplace}{0.917} & \colorbox{secondplace}{0.947} & 0.119 & \colorbox{secondplace}{0.068} & \colorbox{firstplace}{0.045} \\

    BinocularGS~\cite{han2024binocular} & \colorbox{secondplace}{20.71} & \colorbox{thirdplace}{24.31} & \colorbox{thirdplace}{26.70} & \colorbox{thirdplace}{0.862} & \colorbox{secondplace}{0.917} & \colorbox{secondplace}{0.947} & \colorbox{thirdplace}{0.111}& \colorbox{thirdplace}{0.073} & \colorbox{secondplace}{0.052} \\

    NexusGS~\cite{zheng2025nexusgs} & \colorbox{thirdplace}{20.21} & - & - & \colorbox{secondplace}{0.869} & - & - & \colorbox{secondplace}{0.102} & - & - \\

    \midrule
    \textbf{Ours} & \colorbox{firstplace}{\textbf{21.77}} & \colorbox{firstplace}{\textbf{25.09}} & \colorbox{firstplace}{\textbf{27.19}} & \colorbox{firstplace}{\textbf{0.888}} & \colorbox{firstplace}{\textbf{0.928}} & \colorbox{firstplace}{\textbf{0.953}} & \colorbox{firstplace}{\textbf{0.092}} & \colorbox{firstplace}{\textbf{0.064}} & \colorbox{firstplace}{\textbf{0.045}} \\
    \bottomrule
    \end{tabular}
    }
    \vspace{-0.3cm}
  \end{table}

  \begin{figure*}[t]
    \begin{center}
       \includegraphics[trim={0cm 0cm 0cm 0cm},clip,width=0.78\linewidth]{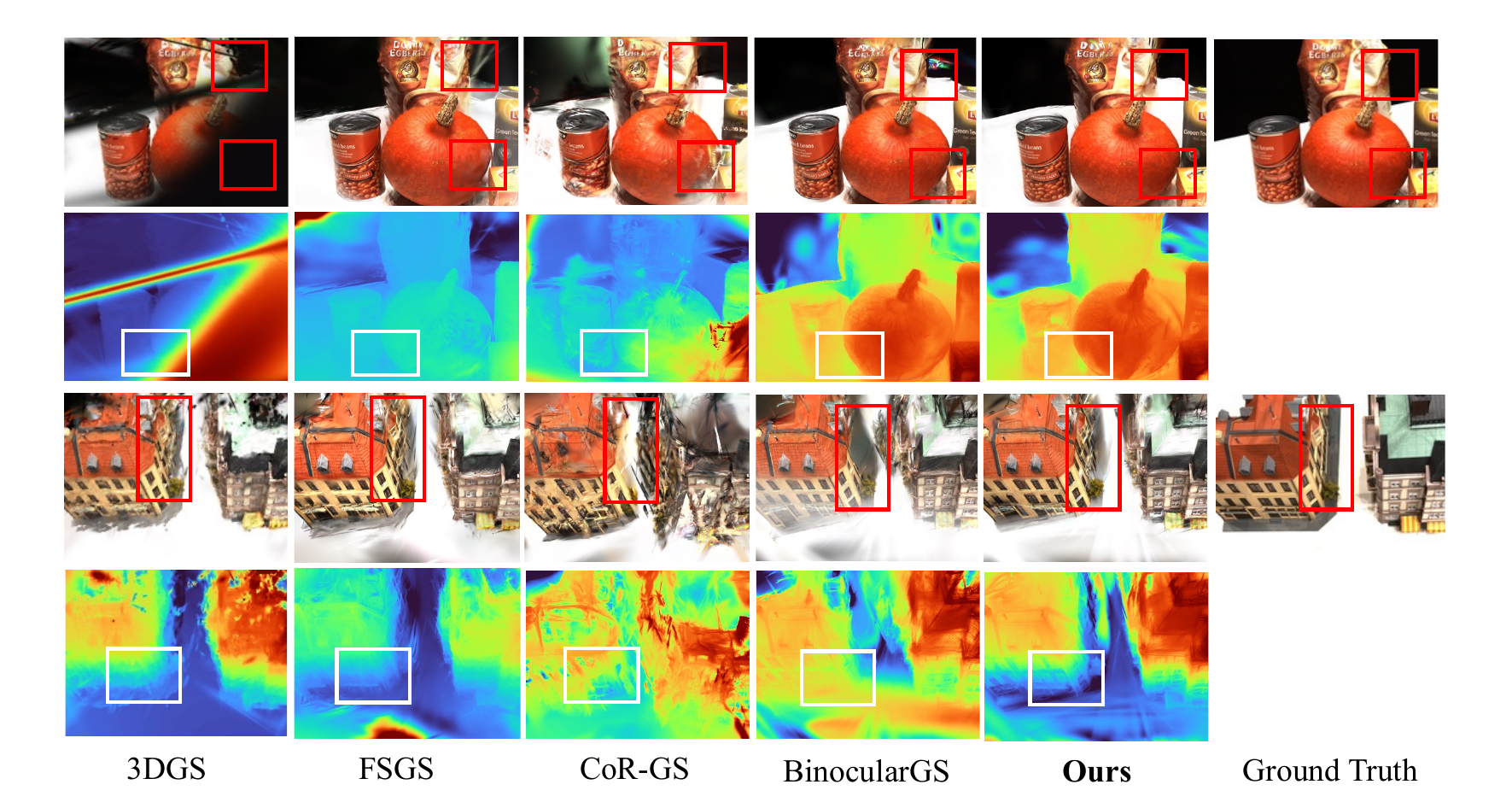}
    \end{center}
    \vspace{-0.6cm}
    \caption{\textbf{Visual comparison on DTU~\cite{jensen2014large} dataset.}}
    \label{fig:dtucom}
    \vspace{-0.3cm}
    \end{figure*}

\subsection{Datasets}
We conduct experiments on three standard benchmarks: LLFF~\cite{mildenhall2019local} 
for forward-facing scenes, DTU~\cite{jensen2014large}, a challenging dataset with 
extensive weakly-textured regions, for object-centric captures,  
and Blender~\cite{mildenhall2021nerf} for 360° object-centric scenes.
Following~\cite{han2024binocular,zhang2024cor,jang2025comapgs}, 
we use 3, 6, 9 training views for LLFF/DTU and 8 views for Blender scenes.
Input images are downsampled by 8× (LLFF), 4× (DTU), and 2× (Blender) to balance 
quality and efficiency, consistent with prior work~\cite{han2024binocular, zhang2024cor,jang2025comapgs}.


\subsection{Implementation Details}
We build upon the BinocularGS framework~\cite{han2024binocular} with dense point cloud 
initialization for LLFF and DTU, and random initialization for Blender.
We train for 30k iterations on LLFF and DTU datasets, and 7k iterations on Blender, 
consistent with the baseline.
Geometric regularization begins at iteration 20k for LLFF/DTU and 4k for 
Blender, while geometry-guided appearance optimization starts at iteration 
25k and 5k respectively. Both operate at every iteration once activated.
All experiments run on the NVIDIA L40s GPU.
Loss weights are set to $\lambda_{\text{mpc}} = 0.1$, 
$\lambda_{\text{smooth}} = 0.01$, and $\lambda_{\text{app}} = 1.0$.
All warping operations are accelerated using batched parallel processing in PyTorch.
We report average results over three independent runs with different seeds.

\begin{table}[t]
    \centering
    \caption{\textbf{Quantitative comparison on Blender~\cite{mildenhall2021nerf} for 8 views.}}
    \label{tab:blender}
    \resizebox{0.60 \columnwidth}{!}{
    \begin{tabular}{l ccc}
    \toprule
    \textbf{Methods} & \textbf{PSNR$\uparrow$} & \textbf{SSIM$\uparrow$} & \textbf{LPIPS$\downarrow$} \\
    \midrule
    DietNeRF~\cite{jain2021dietnerf} & 22.50 & 0.823 & 0.124 \\
    RegNeRF~\cite{niemeyer2022regnerf} & 23.86 & 0.852 & 0.105 \\
    FreeNeRF~\cite{yang2023freenerf} & 24.26 & 0.883 & 0.098 \\
    SparseNeRF~\cite{wang2023sparsenerf} & 22.41 & 0.861 & 0.199 \\
    \midrule
    3DGS~\cite{kerbl20233d} & 22.23 & 0.858 & 0.114 \\
    FSGS~\cite{zhu2024fsgs} & 22.76 & 0.829 & 0.157 \\
    DNGaussian~\cite{li2024dngaussian} & 24.31 & 0.886 & \colorbox{secondplace}{0.088} \\
    CoR-GS~\cite{zhang2024cor} & 23.98 & \colorbox{secondplace}{0.891} & 0.094 \\
    BinocularGS~\cite{han2024binocular} & \colorbox{thirdplace}{24.71} & 0.872 & 0.101 \\
    NexusGS~\cite{zheng2025nexusgs} & 24.37 & \colorbox{firstplace}{0.893 }& \colorbox{firstplace}{0.087} \\
    DropGaussians~\cite{park2025dropgaussian} &\colorbox{secondplace}{25.42} & \colorbox{thirdplace}{0.888} & \colorbox{thirdplace}{0.089} \\
    \midrule
    \textbf{Ours} & \colorbox{firstplace}{\textbf{25.56}} & {\textbf{0.884}} & {\textbf{0.100}} \\
    \bottomrule
    \end{tabular}
    }
    \vspace{-0.3cm}
    \end{table}
  
    \begin{figure}[t]
      \vspace{-0.3cm}
      \begin{center}
         \includegraphics[trim={0cm 0cm 0cm 0cm},clip,width=0.8\linewidth]{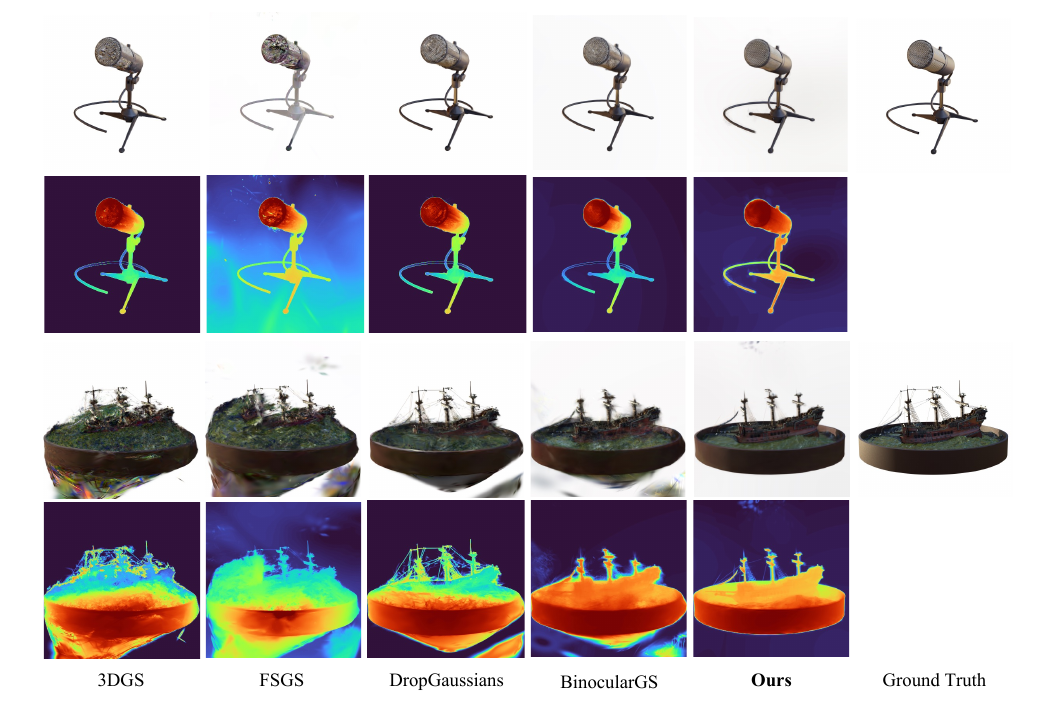}
      \end{center}
      \vspace{-0.3cm}
      \caption{\textbf{Visual comparison on Blender~\cite{mildenhall2021nerf} dataset.}}
      \label{fig:blendercom}
      \vspace{-0.3cm}
      \end{figure}

\subsection{Baselines}
We compare against state-of-the-art methods from both NeRF and 3D Gaussian 
Splatting (3DGS) paradigms.
For NeRF-based methods, we include DietNeRF~\cite{jain2021dietnerf}, 
RegNeRF~\cite{niemeyer2022regnerf}, FreeNeRF~\cite{yang2023freenerf}, 
and SparseNeRF~\cite{wang2023sparsenerf}.
For 3DGS-based methods, we evaluate vanilla 3DGS~\cite{kerbl20233d} and 
sparse-view variants: FSGS~\cite{zhu2024fsgs}, DNGaussian~\cite{li2024dngaussian}, 
CoR-GS~\cite{zhang2024cor}, BinocularGS~\cite{han2024binocular}, 
DropGaussians~\cite{park2025dropgaussian}, NexusGS~\cite{zheng2025nexusgs}, 
and ComapGS~\cite{jang2025comapgs}.

\subsection{Comparisons}
\noindent \textbf{Results on LLFF~\cite{mildenhall2019local} .}
\cref{tab:llff_results} presents quantitative results on the LLFF dataset. 
Our method achieves state-of-the-art performance across all view settings: 
\textbf{+0.76 dB} improvement at 3 views, \textbf{+0.17 dB} over 
ComapGS~\cite{jang2025comapgs} at 6 views, and competitive performance 
at 9 views (\textbf{+0.24 dB} over the third-best). \cref{fig:llffcom} 
demonstrates fewer artifacts in fine details and sharper object boundaries. 
Notably, our depth maps exhibit significantly clearer geometric edges in 
both background textures and foreground structures.

\noindent \textbf{Results on DTU~\cite{jensen2014large}.} 
\cref{tab:dtu_results} shows consistent improvements across all settings: 
\textbf{+1.06 dB}, \textbf{+0.58 dB}, and \textbf{+0.01 dB} over baselines 
at 3, 6, and 9 views, respectively. \cref{fig:dtucom} shows our method 
renders clearer textures in regions (red boxes) via occlusion-aware 
photometric consistency (\cref{subsec:geometry}), while depth maps reveal 
sharper boundaries and finer details (white boxes).

\noindent \textbf{Results on Blender~\cite{mildenhall2021nerf}.} 
For 360° object-centric evaluation, \cref{tab:blender} reports results 
on the Blender dataset~\cite{mildenhall2021nerf} under 8-view setting. 
Our method achieves the best PSNR. While SSIM and LPIPS are slightly lower 
than some methods, this trade-off reflects our prioritization of geometric 
accuracy over perceptual optimization. \cref{fig:blendercom} shows 
fine-grained textures and geometrically accurate structures, 
validating that our geometric regularization (\cref{subsec:geometry}) 
and geometry-guided appearance optimization (\cref{subsec:appearance}) 
effectively preserve structural fidelity critical for 3D reconstruction.

\subsection{Analysis}

\noindent \textbf{Ablation Study.}  
\cref{tab:ablation} validates each component's contribution on LLFF (3 views) and DTU (3 views) by removing: robust multi-view photometric 
consistency Loss \textbf{$\mathcal{L}_{\text{mpc}}^{\text{Fea}}$} (introduced in \cref{eq:mpc_final}), depth smoothness Loss \textbf{$\mathcal{L}_{\text{smooth}} $} (introduced in \cref{eq:smooth}), cycle consistency depth filtering (\textbf{CCDF}, introduced in \cref{eq:depth_mask}), 
and virtual-view photometric consistency Loss \textbf{$\mathcal{L}_{\text{app}}$} (introduced in \cref{eq:aug_loss}).All experiments use BinocularGS~\cite{han2024binocular} as the baseline model.

\begin{figure}[t]
    \vspace{-0.3cm}
    \begin{center}
       \includegraphics[trim={1cm 1cm 1cm 0cm},clip,width=1.0\linewidth]{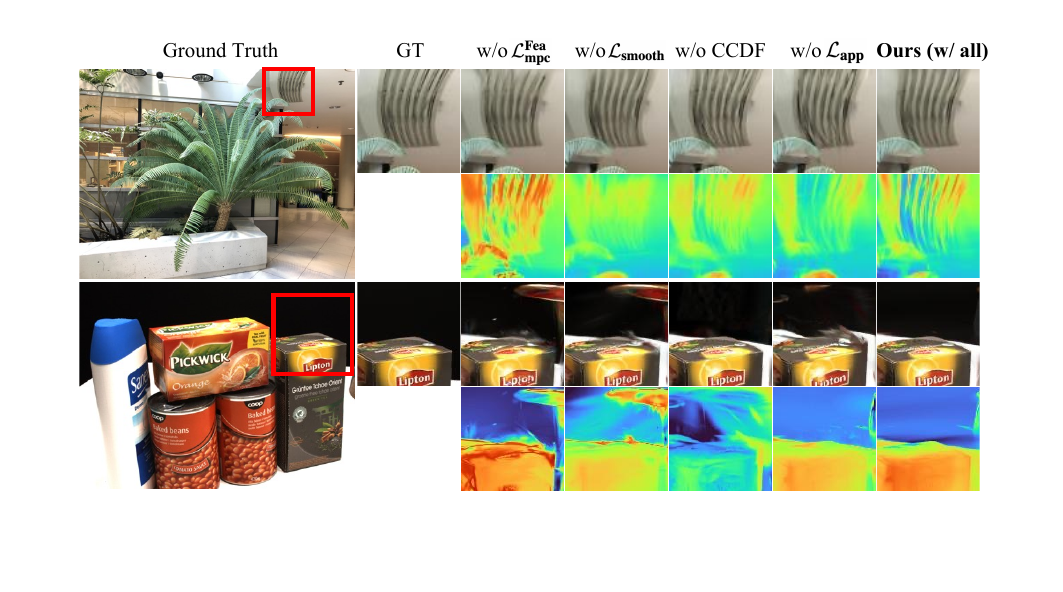}
    \end{center}
    \vspace{-0.8cm}
    \caption{\textbf{Visualization of ablation study results using 3-views.}}
    \label{fig:ablation}
    \vspace{-0.3cm}
    \end{figure}

Removing \textbf{$\mathcal{L}_{\text{mpc}}^{\text{Fea}}$} causes severe degradation 
(LLFF: -0.38 dB, DTU: -0.46 dB). 
As shown in \cref{fig:ablation}, this leads to noticeable blur and noise in both 
RGB and depth, confirming that robust geometric regularization is essential.
\textbf{$\mathcal{L}_{\text{smooth}}$} contributes -0.10 dB on DTU and prevents 
loss of structural information in depth maps (\cref{fig:ablation}), effectively 
mitigating artifacts from depth discontinuities.
\textbf{CCDF} improves virtual view synthesis (DTU: -0.52 dB, 
LLFF: -0.34 dB without it). 
Without CCDF, visible rendering artifacts appear (\cref{fig:ablation}), indicating 
insufficient appearance-geometry consistency.
\textbf{$\mathcal{L}_{\text{app}}$} contributes -0.57 dB on DTU and maintains image 
sharpness (\cref{fig:ablation}), demonstrating that appearance optimization benefits 
from explicit geometric guidance.
Our full model integrates all components to achieve superior geometric fidelity 
with clear structures and crisp boundaries.

\noindent \textbf{Limitations.}  
Our method assumes view-independent appearance during virtual view synthesis. 
In regions with strong view-dependent effects (e.g., specular highlights and 
reflections), the warped appearance may provide incorrect supervision.
However, due to view sparsity, prior methods also struggle in these regions. 
We refer readers to the supplementary material for visual examples.
Despite our accelerated implementation, it requires additional 
computation, resulting in a training time approximately 1.5× that of the baseline. 
The memory overhead is modest (~0.3 GB on 3-view LLFF scenes).
We believe these costs are acceptable given the improvements in 
rendering quality and geometric accuracy under sparse views.

\begin{table}[t]
    \centering
    \caption{\textbf{Ablation studies on LLFF~\cite{mildenhall2019local} and DTU~\cite{jensen2014large}.}}
    \label{tab:ablation}
    \resizebox{1.0\linewidth}{!}{
      \begin{tabular}{cccc|ccc|ccc}
    \toprule
    \multirow{2}{*}{
    $\mathcal{L}_{\text{mpc}}^{\text{Fea}}$
    } & \multirow{2}{*}{
    $\mathcal{L}_{\text{smooth}} $
    } & \multirow{2}{*}{\textbf{CCDF}} & \multirow{2}{*}{
    $\mathcal{L}_{\text{app}}$
    } & \multicolumn{3}{c|}{\textbf{LLFF}(3-view)} & \multicolumn{3}{c}{\textbf{DTU}(3-view)}
     \\

    \cmidrule(lr){5-7} \cmidrule(lr){8-10} 
    & & & & \textbf{PSNR}$\uparrow$ & \textbf{SSIM}$\uparrow$ & \textbf{LPIPS}$\downarrow$ & \textbf{PSNR}$\uparrow$ & \textbf{SSIM}$\uparrow$ & \textbf{LPIPS}$\downarrow$ 
    \\
    \midrule
   $\times$       &$\times$       & $\times$      & $\times$      & {21.44} & {0.751} & {0.168} & {20.71} & {0.862}& {0.111}
     \\

 $\times$       &\checkmark&\checkmark &\checkmark & 21.82 & 0.761 & 0.165 & 21.31 & 0.883 & 0.099 
  \\
  \checkmark& $\times$       &\checkmark &\checkmark& 22.16 & 0.775 & 0.159 & 21.67 & 0.879 & 0.093
   \\

  \checkmark  &  \checkmark  &  $\times$       & \checkmark  & 21.86 & 0.767 & 0.162 & 21.25 & 0.875& 0.104
   \\

  \checkmark  & \checkmark  & \checkmark & $\times$       & 21.79 & 0.763 & 0.164 & 21.20 & 0.870& 0.105
   \\

    \midrule
  \checkmark&\checkmark&\checkmark &\checkmark & {\textbf{22.20}} & {\textbf{0.778}} & {\textbf{0.157}} & {\textbf{21.77}} & {\textbf{0.888}} & {\textbf{0.092}} 
  \\
    \bottomrule
    \end{tabular}
    }
  \end{table}

%% file: secs/5_conc.tex
\section{Conclusion}
\label{sec:conc}

We have presented a novel framework for sparse-view novel view synthesis that effectively addresses the geometry-appearance discrepancy problem in 3D Gaussian Splatting. 
Through robust geometric regularization, geometry-guided appearance optimization, our method achieves joint optimization of geometry and appearance without relying on external depth priors. Extensive experiments demonstrate state-of-the-art performance with reduced floater artifacts and improved geometric accuracy.
